\documentclass[conference]{IEEEtran}
\usepackage{cite}
\usepackage{amsmath,amssymb,amsfonts}
\usepackage{algorithmic}
\usepackage{graphicx}
\usepackage{textcomp}
\usepackage{xcolor}
\usepackage{url}
\usepackage{array}
\usepackage{colortbl}
\usepackage{float}
\def\BibTeX{{\rm B\kern-.05em{\sc i\kern-.025em b}\kern-.08em
    T\kern-.1667em\lower.7ex\hbox{E}\kern-.125emX}}

\begin{document}

\title{Evaluating Stenosis Detection with Grounding DINO, YOLO, and DINO-DETR | ARCADE Dataset}

\author{\IEEEauthorblockN{M6.2: Medical Engineering Research Project | Project Report}
\IEEEauthorblockA{\textit{Medical Engineering - Medical Image and Data Processing} \\
\textit{FAU Erlangen-Nürnberg}\\
Erlangen, Germany \\
Muhammad Musab Ansari \\
muhammad.musab.ansari@fau.de (ib32odij) \\
Matriculation No: 23072475} \\
Supervisors: Sheethal Bhat, Prof. Dr.-Ing. habil. Andreas Maier
}

\maketitle

\begin{abstract}
Detecting stenosis in coronary angiography is vital for diagnosing and managing cardiovascular diseases. This study evaluates the performance of state-of-the-art object detection models on the ARCADE dataset using the MMDetection framework. The models are assessed using COCO evaluation metrics, including Intersection over Union (IoU), Average Precision (AP), and Average Recall (AR). Results indicate variations in detection accuracy across different models, attributed to differences in algorithmic design, transformer-based vs. convolutional architectures. Additionally, several challenges were encountered during implementation, such as compatibility issues between PyTorch, CUDA, and MMDetection, as well as dataset inconsistencies in ARCADE. The findings provide insights into model selection for stenosis detection and highlight areas for further improvement in deep learning-based coronary artery disease diagnosis.
\end{abstract}

\begin{IEEEkeywords}
Object Detection, Stenosis Detection, Deep Learning, Transformer-based Models, Convolutional Neural Networks, COCO Metrics, Medical Imaging
\end{IEEEkeywords}

\section{Introduction}
Coronary artery disease (CAD) is one of the leading causes of mortality worldwide \cite{b1}, necessitating early and accurate detection of stenotic lesions for effective treatment . X-ray coronary angiography (XCA) is the gold standard for visualizing arterial blockages \cite{b2}, but manual interpretation remains labor-intensive and prone to inter-observer variability \cite{b3,b4}. Deep learning-based object detection models have demonstrated significant potential in automating stenosis detection, improving diagnostic efficiency, and reducing clinician workload.

Among the commonly used deep learning architectures, convolutional neural networks (CNNs) such as You Only Look Once (YOLO) \cite{b5} are optimized for real-time inference, making them suitable for applications requiring high-speed detection. On the other hand, transformer-based architectures such as DINO-DETR \cite{b6} and Grounding DINO \cite{b7} leverage self-attention mechanisms for enhanced feature representation and improved generalization, particularly in complex medical imaging scenarios.

This study evaluates the performance of these models on the ARCADE dataset \cite{b8}, a publicly available dataset designed for automatic region-based coronary artery disease diagnostics. Using standard object detection metrics—Intersection over Union (IoU), Average Precision (AP), and Average Recall (AR) \cite{b9}—this research provides a comparative analysis of detection accuracy and model effectiveness for stenosis detection.

To provide additional context, architectural diagrams of the DINO-DETR, Grounding DINO, and YOLO models are included in Appendix~\ref{app:models}.

\section{Motivation}
Coronary artery disease (CAD) is a global health crisis, responsible for millions of deaths annually and placing a significant burden on healthcare systems worldwide \cite{b1}. Early and accurate detection of stenotic lesions is critical for timely intervention and improved patient outcomes. However, current diagnostic methods, such as X-ray coronary angiography (XCA), rely heavily on manual interpretation, which is not only labor-intensive but also prone to inter-observer variability \cite{b3,b4}. This variability can lead to inconsistent diagnoses, potentially delaying treatment and compromising patient care.

While deep learning-based object detection models have shown promise in automating stenosis detection, their adoption in clinical practice remains limited. Existing models, such as YOLO \cite{b5}, prioritize real-time inference but often struggle with the complexity and variability of medical imaging data. On the other hand, transformer-based architectures like DINO-DETR \cite{b6} and Grounding DINO \cite{b7} offer improved feature representation but face challenges in scalability and computational efficiency. These limitations highlight the need for a comprehensive evaluation of state-of-the-art models to identify the most effective approach for stenosis detection.

Moreover, the lack of standardized datasets and evaluation metrics further complicates the development and deployment of automated diagnostic tools. The ARCADE dataset \cite{b8} provides a valuable resource for benchmarking, but inconsistencies in annotations and compatibility issues with deep learning frameworks like MMDetection, PyTorch, and CUDA pose significant technical barriers. Addressing these challenges is essential for bridging the gap between research and real-world clinical applications.

This study aims to address these gaps by providing a systematic evaluation of CNN-based and transformer-based models for stenosis detection. By comparing their performance on the ARCADE dataset using standardized metrics such as Intersection over Union (IoU), Average Precision (AP), and Average Recall (AR) \cite{b9}, this research seeks to identify the most robust and efficient approach for automated CAD diagnostics. The findings are expected to inform the development of clinically viable solutions, ultimately improving diagnostic accuracy and reducing the burden on healthcare providers.

\section{Dataset}
The ARCADE dataset (Automatic Region-based Coronary Artery Disease Diagnostics using X-ray Angiography Images) is a publicly available benchmark dataset designed to facilitate the development and evaluation of automated methods for coronary artery disease (CAD) diagnostics \cite{b8}. It was introduced as part of the ARCADE challenge at the 26th International Conference on Medical Image Computing and Computer-Assisted Intervention (MICCAI). The dataset provides expert-labeled X-ray coronary angiography (XCA) images, enabling researchers to develop and assess deep learning models for vessel segmentation and stenosis detection.

\subsection{Dataset Composition}
The ARCADE dataset consists of two separate tasks, each containing 1500 images:

\begin{itemize}
    \item \textbf{Coronary Vessel Classification:} Images are annotated following the SYNTAX Score methodology, which divides the heart into distinct vascular regions.
    \item \textbf{Stenosis Detection:} Images include annotations of atherosclerotic plaques, marking regions of arterial narrowing.
\end{itemize}

\begin{figure}[htbp]
    \centering
    \includegraphics[width=0.3\linewidth]{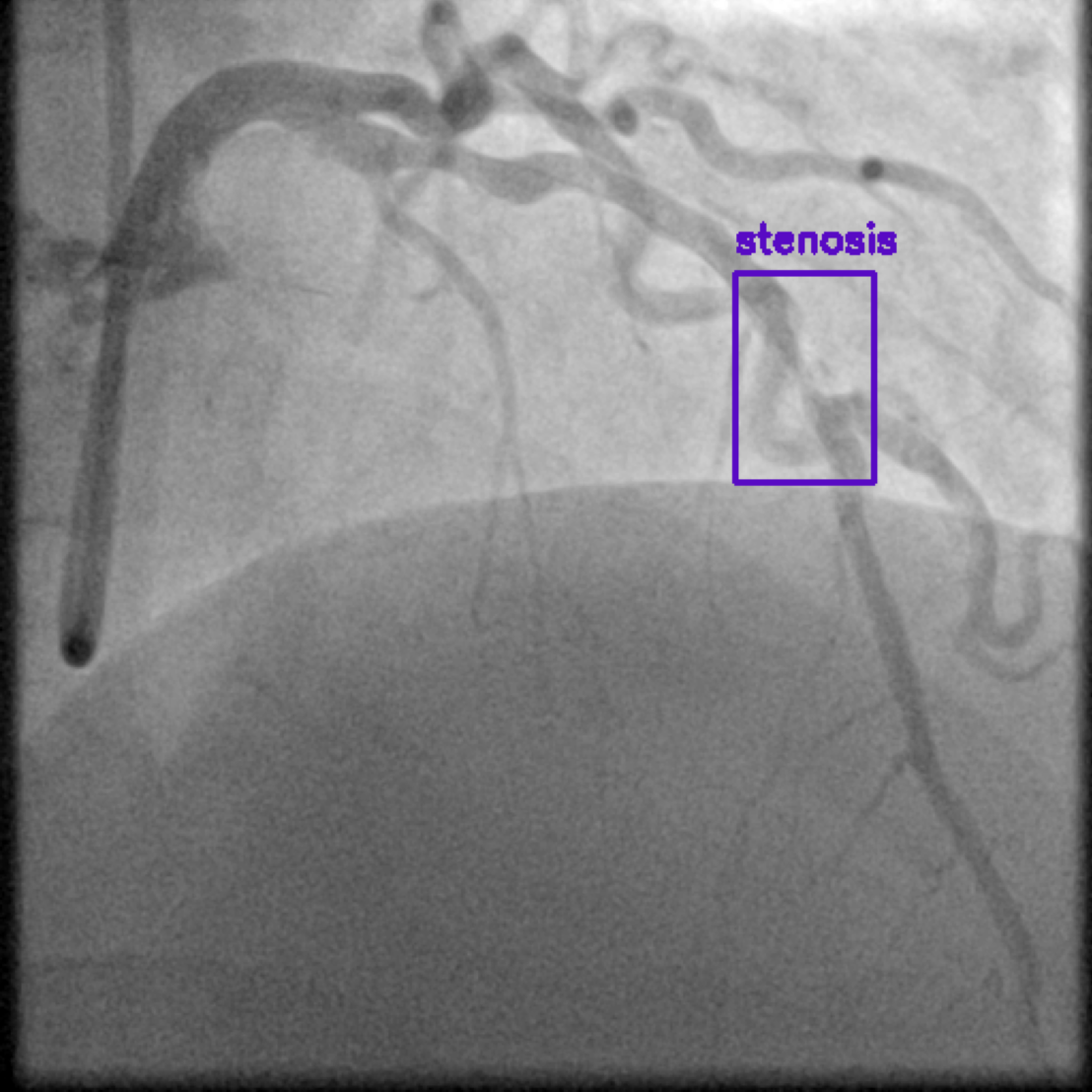}
    \hfill
    \includegraphics[width=0.3\linewidth]{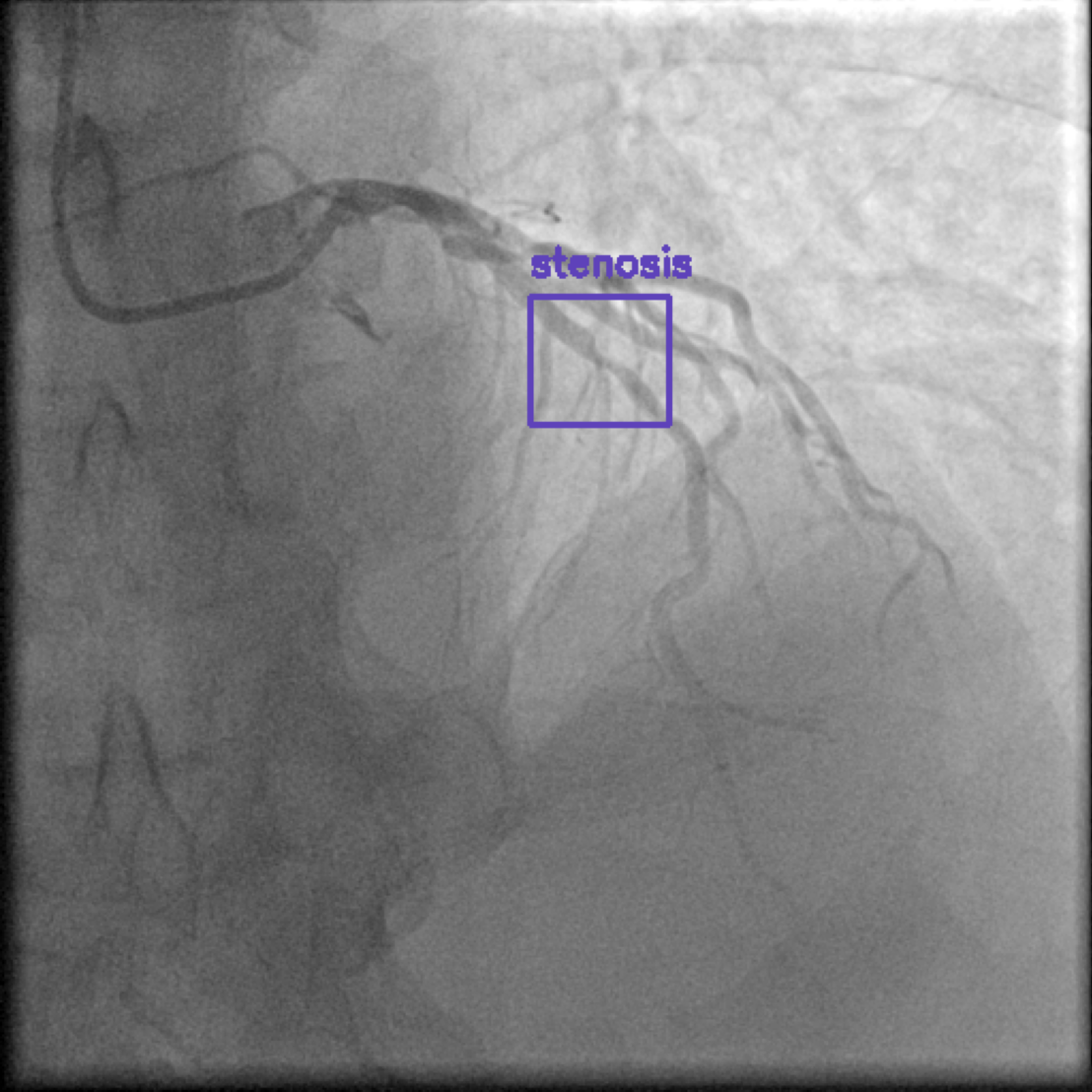}
    \hfill
    \includegraphics[width=0.3\linewidth]{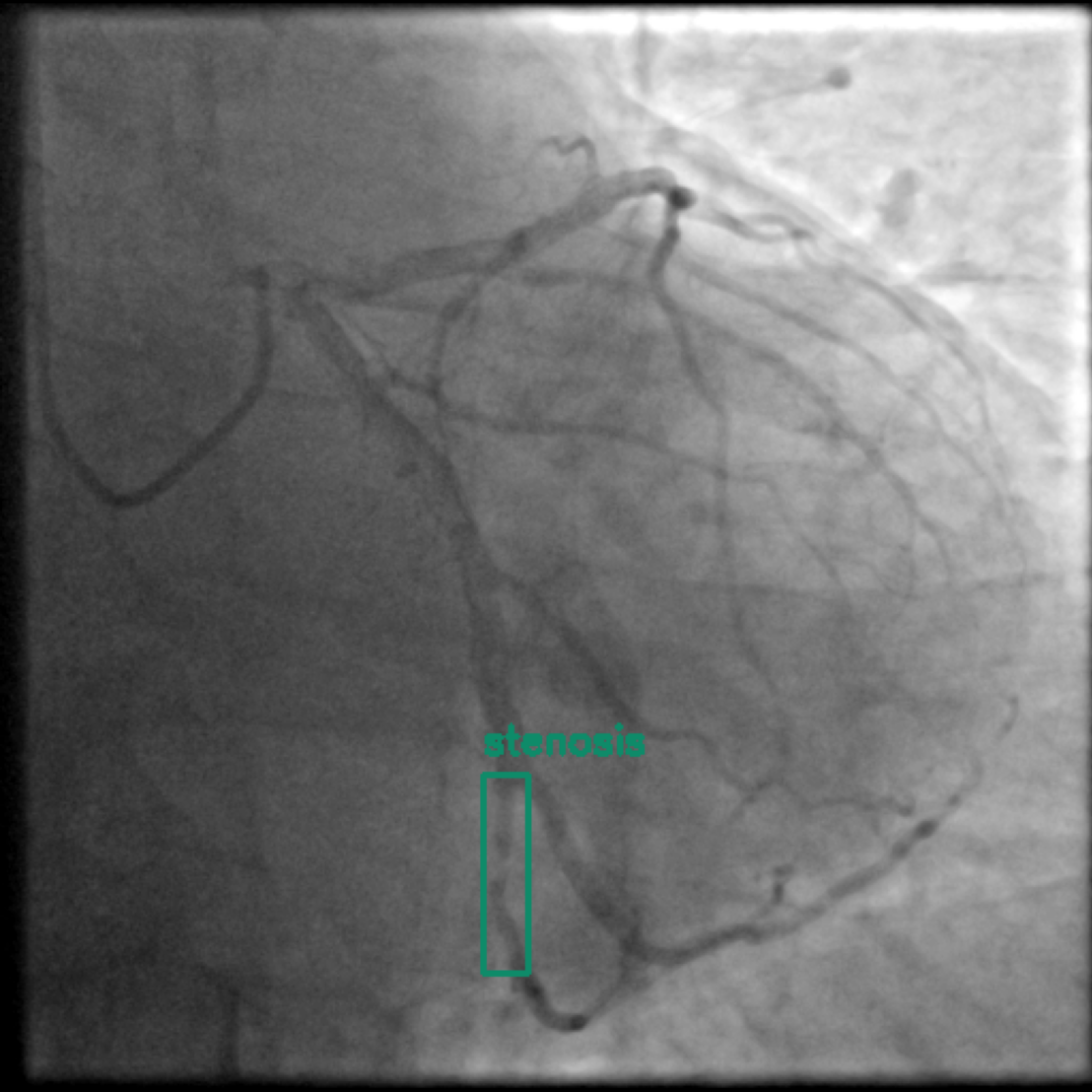}
    \caption{Sample images from the ARCADE dataset with annotations for stenosis detection. The red bounding boxes indicate regions of arterial narrowing.}
    \label{fig:stenosis_annotations}
\end{figure}

Figure~\ref{fig:stenosis_annotations} shows sample images from the ARCADE dataset with annotations for stenosis detection. The annotations highlight regions of arterial narrowing, providing ground truth data for training and evaluating object detection models.

For both tasks, images are categorized into training, validation, and test sets. Annotations are structured in the COCO format, encompassing essential details such as bounding boxes and segmentation masks, and are systematically organized within .JSON files to facilitate model development and evaluation. The COCO format proved particularly useful for DINO-DETR and Grounding DINO architectures. However, necessary adjustments, such as the inclusion of \texttt{data.yaml} and other requirements specific to YOLO, had to be made to accommodate YOLO’s framework.

\subsection{Annotation Process}
The dataset was annotated by medical experts to ensure accurate delineation of vessel regions and stenotic plaques. The annotation process involved identifying coronary artery segments and marking stenotic areas, providing valuable ground truth data for model training and evaluation. However, inconsistencies in the dataset, such as variations in annotation formats and category mismatches, posed challenges in implementation, requiring careful preprocessing before model training.

\subsection{Annotation Format}
The .JSON annotation files consist of the following fields:
\begin{itemize}
    \item \textbf{images:} Contains unique image IDs, dimensions, and file names.
    \item \textbf{categories:} Lists unique IDs and names, corresponding to SYNTAX descriptions.
    \item \textbf{annotations:} Includes annotation IDs, linked image IDs, category IDs, segmentation coordinates in XYXY format, bounding box coordinates in XYWH format, and bounding box area.
\end{itemize}

\subsection{Usage in This Study}
In this study, the stenosis detection task is implemented using the provided bounding box annotations to train and evaluate object detection models. The dataset enables a comparative analysis of convolutional and transformer-based architectures in their capability to detect and localize stenotic regions in XCA images. Performance evaluation is conducted using COCO metrics, including Intersection over Union (IoU), Average Precision (AP), and Average Recall (AR), which provide quantitative insights into model effectiveness for automated CAD diagnosis \cite{b9}.

\section{Literature Review}

The automated detection of coronary artery disease (CAD) in X-ray coronary angiography (XCA) has been extensively studied using deep learning-based methodologies. Various approaches have been proposed to address challenges related to stenosis detection, vessel segmentation, and lesion classification. Object detection models, including convolutional neural networks (CNNs) and transformer-based architectures, have demonstrated promising results in medical image analysis.

\subsection{Deep Learning for Stenosis Detection and Coronary Segmentation}

Several studies have focused on deep learning-based segmentation of coronary arteries, which plays a crucial role in stenosis detection. Zhang et al. \cite{b10} introduced a progressive perception learning (PPL) framework that enhances segmentation accuracy by incorporating context, interference, and boundary perception modules. This approach achieved a Dice score exceeding 95\% on a dataset of 1,086 subjects, outperforming multiple state-of-the-art methods. Similarly, Fazlali et al. \cite{b11} developed a segmentation framework based on superpixels and vesselness probability measures, eliminating the need for labeled training data. Their method demonstrated superior segmentation performance while reducing false positives and computational time compared to traditional graph-cut-based algorithms.

Stenosis detection has also been explored using deep learning-based classification and object detection models. Du et al. \cite{b12} utilized a dataset of 20,612 angiograms to train a deep learning pipeline for segment recognition and lesion classification. Their model achieved high recognition accuracy (98.4\%) and lesion classification F1 scores ranging from 0.802 to 0.854. However, a common limitation among such studies is the unavailability of training data, restricting reproducibility and further advancements.

\subsection{Object Detection in Medical Imaging}

Object detection architectures, such as YOLO \cite{b5}, DINO-DETR \cite{b6}, and Grounding DINO \cite{b7}, have been widely applied in medical image analysis, including lesion detection and anatomical structure segmentation. YOLO, a CNN-based model, has been favored for its real-time processing capabilities, whereas transformer-based models like DINO-DETR and Grounding DINO leverage self-attention mechanisms for improved feature representation and object localization. Recent advancements in DINO-based architectures have demonstrated improved performance in complex detection tasks by leveraging knowledge distillation techniques and query-based object localization. For a detailed visualization of the architectures of YOLO, DINO-DETR, and Grounding DINO, refer to Appendix~\ref{app:models}.

MMDetection \cite{b13}, an open-source toolbox for object detection, has provided a standardized implementation for multiple state-of-the-art models, including YOLO and DINO-based architectures. This framework has been extensively used in medical image analysis due to its modular design and support for various training and evaluation strategies. Studies have demonstrated the effectiveness of MMDetection-based pipelines in anatomical structure detection and disease classification, enabling systematic comparison of different object detection models. The integration of MMDetection within the CAD diagnosis workflow allows for reproducible experiments and efficient hyperparameter tuning, contributing to improved performance in automated stenosis detection.

\subsection{The ARCADE Dataset for CAD Diagnostics}

The ARCADE dataset \cite{b8} provides a publicly available benchmark for automated CAD diagnosis using XCA, containing 1,500 annotated images. Unlike many existing datasets, which remain proprietary or lack detailed annotations, ARCADE includes structured labels for coronary vessel regions and stenotic plaques, allowing for comprehensive analysis of CAD severity. In addition to traditional segmentation and classification tasks, this dataset enables the evaluation of region-based object detection models for stenosis assessment.

By facilitating comparative analysis of YOLO, DINO-DETR, and Grounding DINO, the ARCADE dataset contributes to advancing the development of automated CAD diagnostic systems. The availability of a standardized dataset aims to bridge the gap in reproducibility and support further research in the field. Furthermore, physics-guided deep learning methods have been explored to improve cardiovascular disease assessment in smart healthcare applications \cite{b14}, demonstrating the potential of integrating deep learning with domain-specific knowledge for enhanced diagnostic accuracy.

\section{Evaluation Metrics}\label{evaluationmetrics}

The performance of object detection models on the ARCADE dataset was assessed using COCO evaluation metrics, including Average Precision (AP) and Average Recall (AR) across various Intersection over Union (IoU) thresholds and object scales \cite{b9}. These metrics provide a comprehensive view of each model's detection capabilities and allow for comparative analysis (see Table~\ref{tab:evaluation_metrics}).

\begin{table}[ht]
\caption{Model Evaluation Metrics on the ARCADE Dataset}
\centering
\begin{tabular}{|l|c|c|c|}
\hline
\textbf{Metric} & \textbf{DINO-DETR} & \textbf{YOLO} & \textbf{Grounding DINO} \\
\hline
mAP@[0.50:0.95] & 0.086 & 0.068 & 0.080 \\
mAP50 & 0.228 & 0.254 & 0.259 \\
mAP75 & 0.056 & 0.019 & 0.034 \\
mAP (small) & 0.198 & 0.126 & 0.168 \\
AR@[0.50:0.95] (100) & 0.526 & 0.180 & 0.416 \\
AR@[0.50:0.95] (300) & 0.621 & 0.180 & 0.469 \\
AR@[0.50:0.95] (1000) & 0.621 & 0.180 & 0.469 \\
AR (small) & 0.548 & 0.148 & 0.413 \\
AR (medium) & 0.734 & 0.229 & 0.555 \\
\hline
\end{tabular}
\label{tab:evaluation_metrics}
\end{table}

\subsection{Metric Descriptions}

\textbf{Mean Average Precision (mAP):} mAP measures the accuracy of a model by averaging the precision over multiple IoU thresholds (e.g., 0.50 to 0.95) \cite{b9}. The higher the mAP, the better the model's precision and its ability to localize and classify objects correctly. This metric reflects the model's performance over varying levels of overlap between predicted and true bounding boxes.

\textbf{mAP@50 and mAP@75:} These are specific instances of mAP evaluated at singular IoU thresholds of 0.50 and 0.75, respectively \cite{b9}. mAP@50 is often used as a baseline due to its allowance for a larger margin of prediction error, while mAP@75 represents stricter criteria for successful detection with higher overlap required.

\textbf{Average Recall (AR):} AR assesses how well a model retrieves relevant instances of an object, averaged over multiple IoU thresholds \cite{b9}. Higher AR values indicate better performance in capturing true positives, even at higher maximum detection levels.

\textbf{Maximum Detections (100, 300, 1000):} The numbers in parentheses following AR (e.g., AR@[0.50:0.95] (100)) indicate the maximum number of detections considered per image during evaluation \cite{b9}:
\begin{itemize}
    \item \textbf{(100):} Measures AR when a maximum of 100 detections per image is allowed, useful for evaluating models in settings with typically fewer detectable objects.
    \item \textbf{(300):} Measures AR with up to 300 detections, offering a balance between speed and the ability to capture multiple instances.
    \item \textbf{(1000):} Allows up to 1000 detections per image, providing insights into a model's capacity to manage large numbers of detections, as might be necessary for densely populated scenes.
\end{itemize}

\textbf{Scale-specific Metrics (mAP small, AR small, AR medium):} These metrics focus on model performance across different object scales \cite{b9}. mAP and AR for small objects measure how effectively small instances are detected, which is crucial for applications where fine-grained detection is needed, such as medical imaging. AR (medium) provides insights into the model's ability to detect medium-sized objects.

\section{Performance Analysis}

The evaluation metrics presented in Table~\ref{tab:evaluation_metrics} offer valuable insights into the performance of different object detection models on the ARCADE dataset. This section analyzes the results, focusing on the comparative strengths and weaknesses of each model in the context of coronary artery disease (CAD) detection using X-ray coronary angiography (XCA).

\subsection{Model Precision and Recall}

Transformer-based models, such as DINO-DETR and Grounding DINO, demonstrate a generally higher mean Average Precision (mAP) across most IoU thresholds compared to YOLO \cite{b9}. Particularly, mAP50 results indicate that these models are more precise in detecting and locating stenotic regions, which may be attributed to their robust feature representation facilitated by self-attention mechanisms \cite{b6,b7}.

Despite YOLO's real-time processing advantage, it exhibits lower precision and recall, as seen in its reduced mAP and AR values \cite{b5}. This trade-off suggests that while YOLO is beneficial for applications requiring faster inference, its accuracy may be compromised in scenarios necessitating detailed image analysis, such as medical diagnostics.

\subsection{Scalability and Object Size Detection}

The comparison of mAP and AR for small and medium objects reveals that DINO-DETR achieves superior performance in detecting smaller-scale stenotic lesions, which are critical in medical imaging contexts \cite{b6}. The ability of DINO-DETR and Grounding DINO to model complex relationships across spatial resolutions allows them to excel over YOLO, particularly in tasks involving subtle and fine-grained features.

\subsection{Detection Capacity}

Analysis of AR across varying maximum detections (100, 300, 1000) illuminates each model's capacity to handle different detection densities within images. DINO-DETR maintains consistent recall performance with increasing detections, suggesting its robustness in richly populated scenes. Conversely, YOLO's recall plateau across detection thresholds underscores a potential limitation in capturing multiple instances, further highlighting its design trade-offs \cite{b5}.

\subsection{Implications for CAD Detection}

The enhanced precision and recall of transformer-based architectures underscore their suitability for CAD detection in XCA images, where accurate delineation of stenotic lesions is critical \cite{b6,b7}. The results advocate for the integration of these advanced models in clinical workflows, promoting higher diagnostic accuracy in automated CAD systems.

This performance analysis highlights the trade-offs associated with each model architecture, guiding optimal model selection for targeted diagnostic applications in coronary artery disease detection.

\section{Results}

\subsection{Quantitative Results}
The performance of the models was evaluated using the ARCADE dataset. The results, including mean Average Precision (mAP) and Average Recall (AR) metrics, are summarized in Table~\ref{tab:evaluation_metrics} (refer to Section~\ref{evaluationmetrics} for details). 

The evaluation metrics presented in Table~\ref{tab:evaluation_metrics} illustrate the differences in model performance. Grounding DINO achieved the highest mAP at IoU = 0.50, demonstrating its precision in detecting and localizing objects with moderate overlap \cite{b7}. DINO-DETR outperformed the other models in mAP across IoU thresholds from 0.50 to 0.95, indicating its effectiveness in consistently capturing objects of varying overlap levels \cite{b6}. YOLO, while excelling in real-time processing speed, showed competitive mAP50 results, reflecting a balanced performance in precision for moderately overlapping objects \cite{b5}.

\subsection{Qualitative Results}

\begin{figure*}[!htbp]
    \centering
    \renewcommand{\arraystretch}{1.3}
    \setlength{\tabcolsep}{2pt} 

    \begin{tabular}{cccc}
        \includegraphics[width=0.22\linewidth]{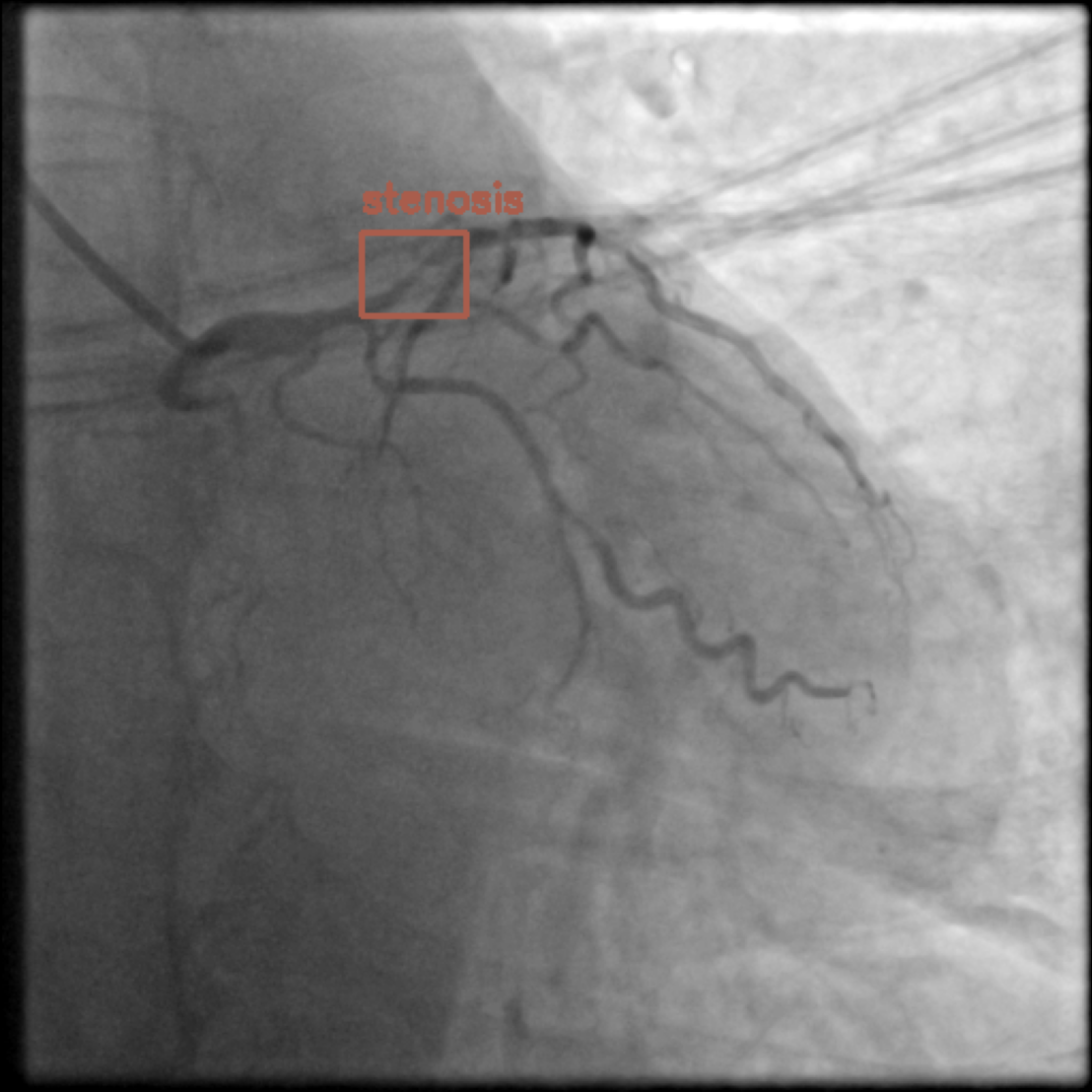} &
        \includegraphics[width=0.22\linewidth]{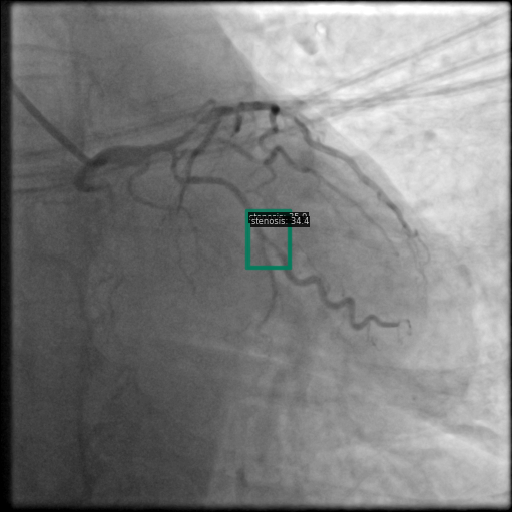} &
        \includegraphics[width=0.22\linewidth]{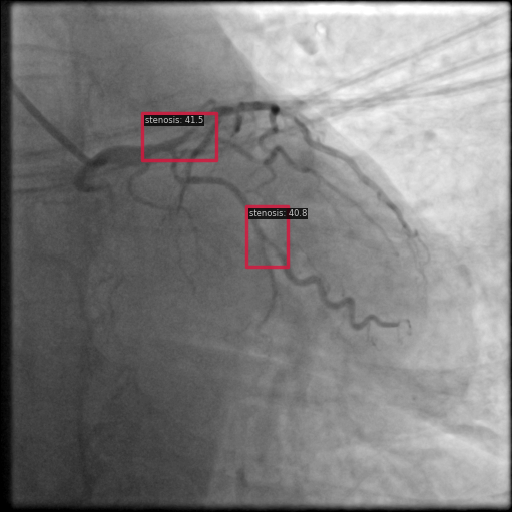} &
        \includegraphics[width=0.22\linewidth]{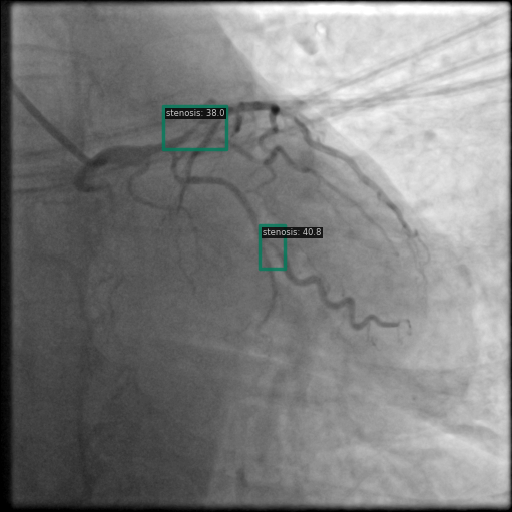}  \\

        \includegraphics[width=0.22\linewidth]{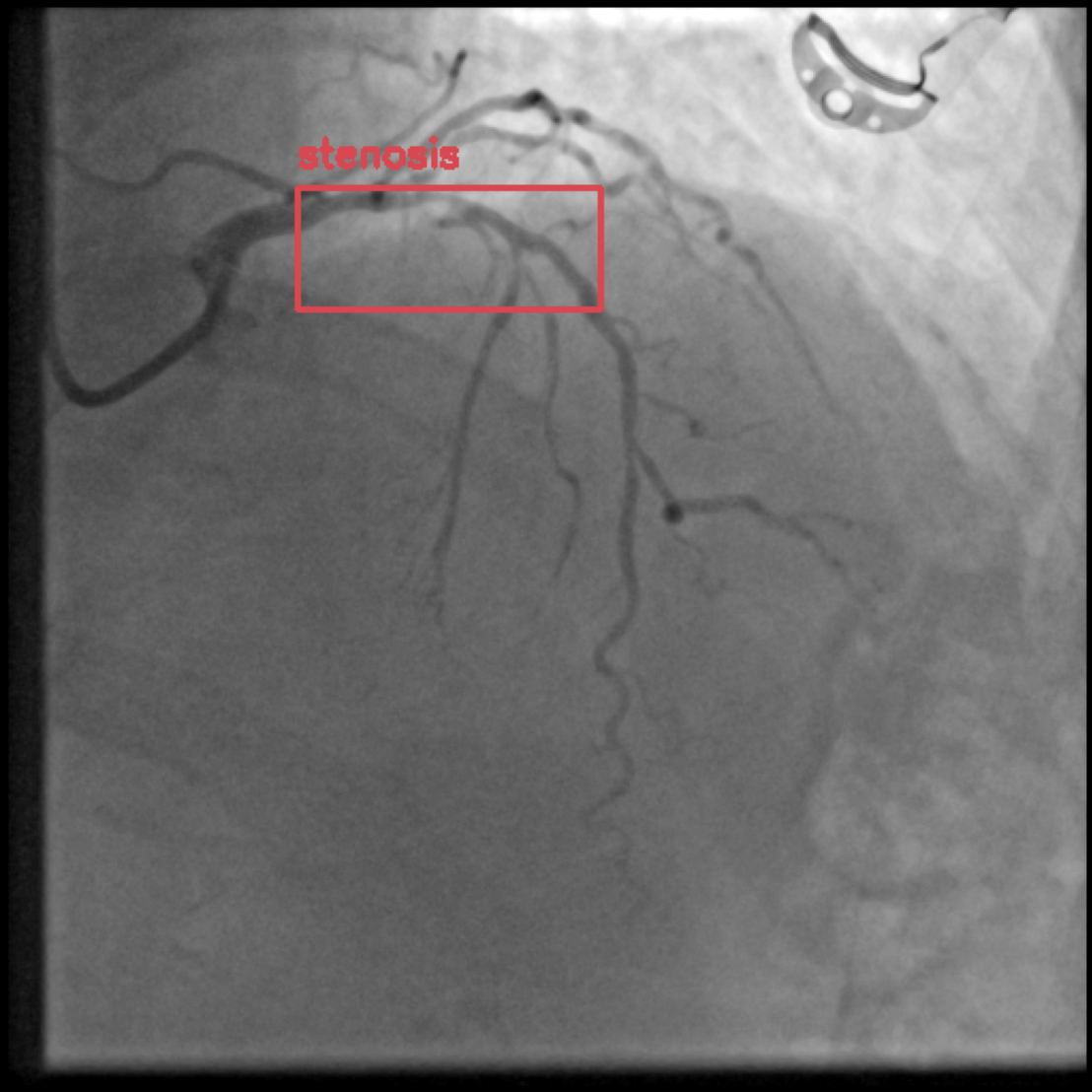} &
        \includegraphics[width=0.22\linewidth]{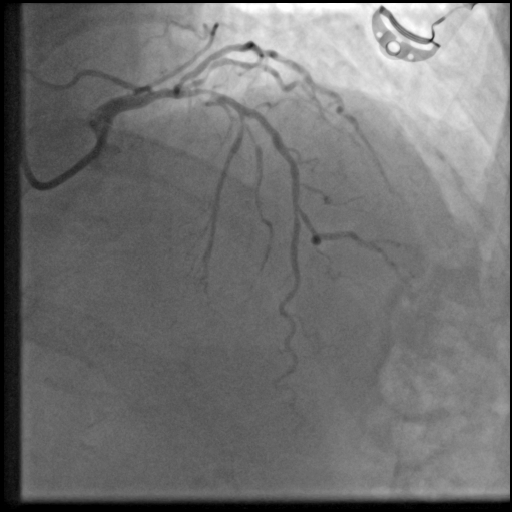} &
        \includegraphics[width=0.22\linewidth]{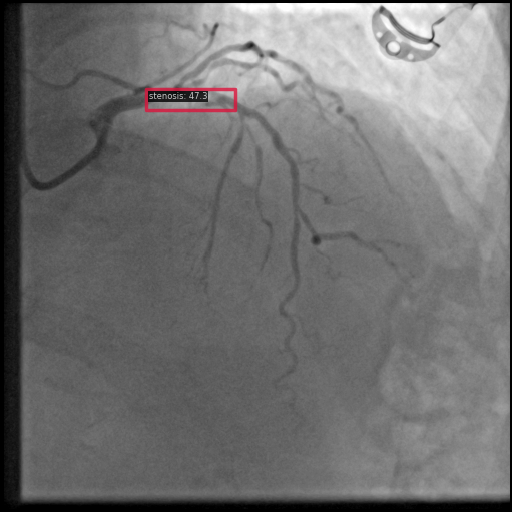} &
        \includegraphics[width=0.22\linewidth]{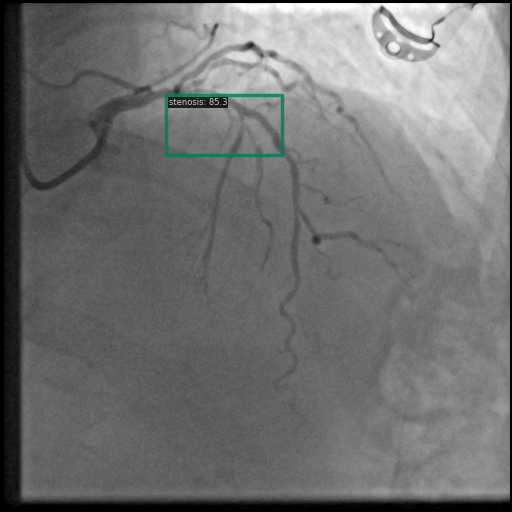} \\

        \includegraphics[width=0.22\linewidth]{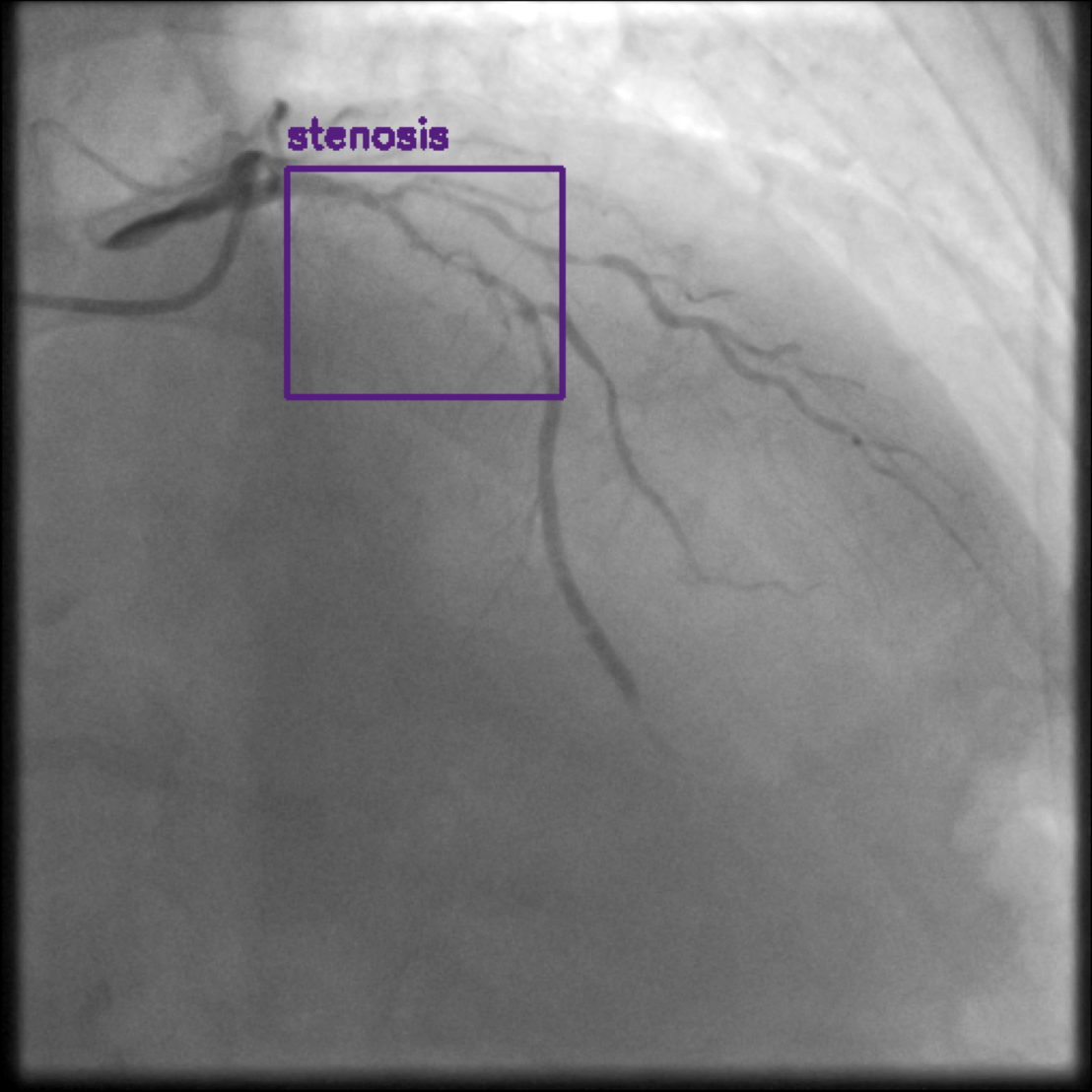} &
        \includegraphics[width=0.22\linewidth]{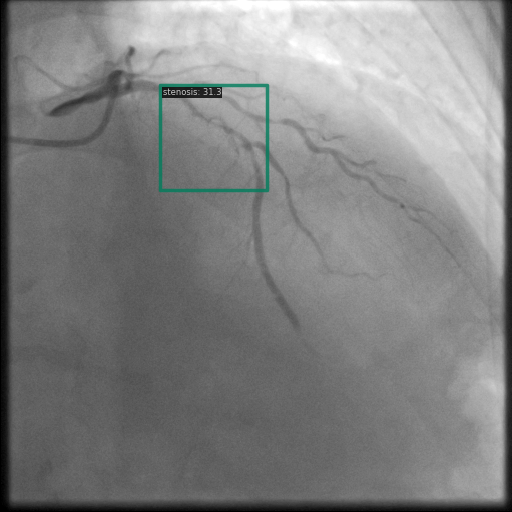} &
        \includegraphics[width=0.22\linewidth]{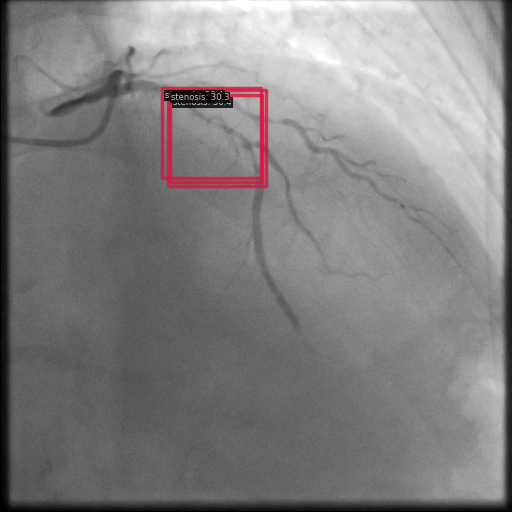} &
        \includegraphics[width=0.22\linewidth]{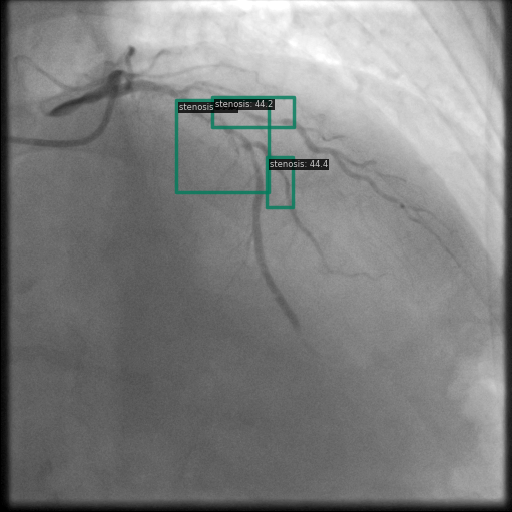} \\

        \includegraphics[width=0.22\linewidth]{3.png} &
        \includegraphics[width=0.22\linewidth]{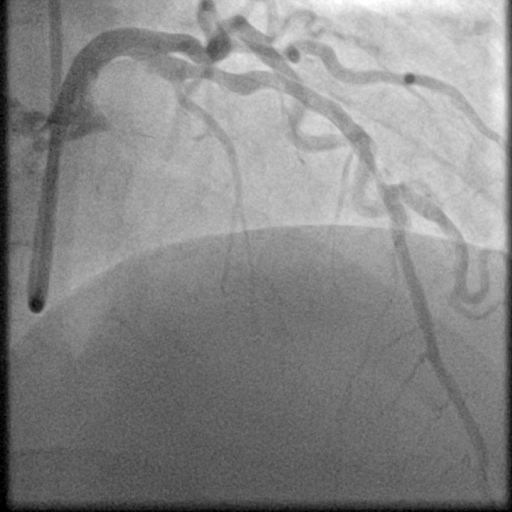} &
        \includegraphics[width=0.22\linewidth]{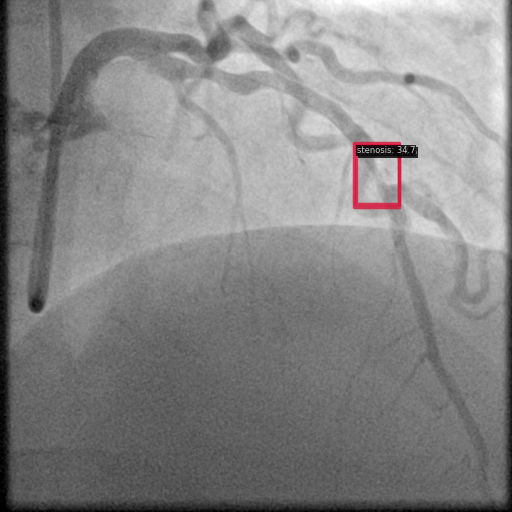} &
        \includegraphics[width=0.22\linewidth]{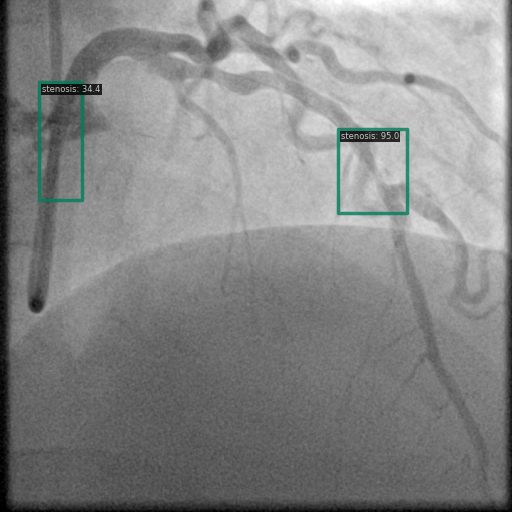} \\
    \end{tabular}

    \vspace{3pt} 
    (a) Original \hspace{50pt} (b) DINO-DETR \hspace{30pt} (c) Grounding DINO \hspace{40pt} (d) YOLO \hspace{10pt}

    \caption{Qualitative comparison of original test images (first column) with detection results from DINO-DETR, Grounding DINO, and YOLO.}
    \label{fig:qualitative_comparison}
\end{figure*}

To further assess the detection performance, qualitative results for three test images are presented in Figure~\ref{fig:qualitative_comparison}. The first column shows the original images with ground truth annotations. The second, third, and fourth columns depict detections from DINO-DETR, Grounding DINO, and YOLO, respectively.

The qualitative comparison in Figure~\ref{fig:qualitative_comparison} reinforces the trends observed in the quantitative evaluation. DINO-DETR produced fewer detections, occasionally missing relevant stenotic regions, consistent with its conservative detection strategy aimed at minimizing false positives \cite{b6}. Grounding DINO, although capable of identifying more regions, sometimes resulted in cluttered predictions due to over-detections \cite{b7}. YOLO provided a reasonable trade-off by effectively capturing anatomical structures while maintaining relatively high confidence scores and clear visualizations \cite{b5}.

\section{Discussion}

The detection performance of DINO-DETR, Grounding DINO, and YOLO was evaluated using both quantitative metrics and qualitative observations. This section analyzes the key findings, discusses the strengths and limitations of each approach, and explores potential post-processing techniques to enhance detection accuracy.

The models were evaluated using their respective configurations from the MMDetection repository: DINO-DETR (dino-5scale\_r50\_4xb2-12e\_coco.py), Grounding DINO (grounding\_dino\_r50\_4xb2-12e\_coco.py), and YOLO (yolov3\_d53\_8xb8-ms-608-273e\_coco.py). These configurations influence detection performance by dictating anchor sizes, learning rates, and augmentation strategies \cite{b13}. Exploring alternative configurations, such as deeper backbones or multi-scale feature fusion techniques, could further optimize detection accuracy.

To refine model predictions and mitigate false positives, several post-processing techniques can be applied. Non-Maximum Suppression (NMS) helps reduce overlapping bounding boxes, while confidence threshold tuning can filter out uncertain predictions while retaining high-confidence ones. Ensemble methods, which combine predictions from multiple models, may leverage their respective strengths to improve robustness \cite{b15}. Additionally, post-hoc calibration techniques like isotonic regression or temperature scaling can adjust confidence scores for better interpretability \cite{b16}.

The current implementation does not include these post-processing techniques due to time constraints. However, as a next step, implementing NMS and confidence threshold tuning for YOLO and DETR models could help filter overlapping or low-confidence predictions. Additionally, integrating an ensemble approach using DETR and YOLO results might improve detection robustness, particularly for challenging cases with faint stenotic regions.

Regarding model configurations, exploring alternative architectures could enhance detection accuracy. Specifically, testing a deeper transformer-based backbone such as Swin Transformer for DINO-DETR or integrating deformable attention layers in Grounding DINO might improve feature extraction and multi-scale representation \cite{b6,b7}. Additionally, applying feature pyramid networks (FPN) in YOLO could enhance its ability to detect small and complex structures within coronary angiography images \cite{b17}. Longer training schedules or larger batch sizes within MMDetection could also be explored to improve model convergence and generalization \cite{b13}.

Beyond architecture and training adjustments, domain-specific augmentation techniques such as vessel-enhancement preprocessing could improve model sensitivity to stenotic regions. Moreover, leveraging semi-supervised learning techniques may be beneficial for enhancing performance when labeled data is limited. Future work could also assess the effectiveness of recent transformer-based object detectors for medical image analysis, determining their suitability for coronary angiography applications.

Beyond architecture and training adjustments, domain-specific augmentation techniques such as vessel-enhancement preprocessing could improve model sensitivity to stenotic regions \cite{b18}. Moreover, leveraging semi-supervised learning techniques may be beneficial for enhancing performance when labeled data is limited \cite{b19}. Future work could also assess the effectiveness of recent transformer-based object detectors for medical image analysis, determining their suitability for coronary angiography applications \cite{b20}.

\subsection{Challenges and Implementation Issues}

During the implementation of object detection models using MMDetection, several challenges were encountered, ranging from model-specific compatibility issues to dataset inconsistencies and computational limitations.

The implementation of all three models from the MMDetection repository—DINO-DETR, YOLO, and Grounding DINO—posed challenges during the initial setup due to library dependencies and version mismatches \cite{b13}. These models required specific versions of MMDetection, MMCV, and PyTorch, leading to compatibility issues across different software environments. Additionally, configurations needed adjustments to align with the ARCADE dataset format, particularly in handling annotation files and ensuring proper data loading pipelines.

The software stack required careful management, as conflicts arose due to CUDA compatibility problems, version mismatches in MMCV and MMDetection, and PyTorch inconsistencies affecting GPU utilization and training efficiency. Resolving these issues involved extensive testing and environment isolation to ensure stable and reproducible training workflows.

Transitioning from a local machine to the High-Performance Computing (HPC) cluster introduced additional challenges. Storage quota issues emerged due to large model checkpoints and environment dependencies, necessitating careful cleanup and efficient storage management. Efficient Slurm job scheduling was critical to optimizing multi-GPU training while avoiding resource wastage. Additionally, remote debugging on the HPC proved more difficult than on a local machine due to limited interactive access, complicating the diagnosis of training failures.

The ARCADE dataset contained inconsistencies, particularly duplicate entries in the train.json annotation file, which led to training instabilities. Pre-processing steps were required to filter redundant annotations, ensuring label consistency across train.json, val.json, and test.json to prevent category mismatches that could impact training and evaluation. Addressing these dataset issues was essential for maintaining reliable model performance.

\subsection{Recommendations}

To address the challenges encountered during the implementation of object detection models using MMDetection, several strategic recommendations are proposed. Effective version management is crucial to help maintain consistent environments. Detailed documentation of installation procedures and version requirements is advised to enhance setup efficiency and troubleshooting. Compatibility testing and the development of automated setup scripts can preemptively resolve conflicts in software stacks, ensuring seamless integration of components like CUDA, MMCV, and PyTorch. In high-performance computing (HPC) environments, optimizing resource allocation and implementing robust data management practices can alleviate storage and efficiency constraints. Enhanced remote debugging tools are recommended to simplify error diagnosis on HPC systems. For datasets, automated scripts to eliminate duplicate entries and rigorous consistency checks across partitions will improve data integrity and minimize training inconsistencies. Implementing these recommendations will streamline future workflows, improve model deployment, and facilitate the reliable integration of advanced detection models in various applications.

\section{Conclusion}

This study evaluated the performance of DINO-DETR, Grounding DINO, and YOLO for stenosis detection in coronary angiography images. While DETR-based models demonstrated strong interpretability and attention-based localization \cite{b20}, YOLO provided faster inference times with competitive accuracy \cite{b21}. Post-processing techniques such as Non-Maximum Suppression (NMS) and confidence threshold tuning were identified as potential strategies to reduce false positives and refine model predictions. However, these were not implemented due to time constraints and remain an area for future exploration.

To further enhance detection accuracy, alternative configurations—such as deeper transformer-based backbones for DETR, deformable attention layers for Grounding DINO \cite{b22}, and feature pyramid networks for YOLO \cite{b23}—could be investigated. Additionally, domain-specific augmentation techniques like vessel-enhancement preprocessing and semi-supervised learning approaches \cite{b19} may improve model sensitivity, especially when labeled data is limited. Tools like Albumentations \cite{b24} and U-Net-inspired architectures \cite{b25} could further enhance preprocessing and feature extraction pipelines.

Despite computational and dataset-related challenges, this work demonstrates the feasibility of applying transformer-based and CNN-based object detection models for medical imaging. Future research should focus on integrating hybrid detection architectures, optimizing training strategies, and leveraging advanced augmentation techniques to improve robustness in real-world clinical settings.

Despite computational and dataset-related challenges, this work demonstrates the feasibility of applying transformer-based and CNN-based object detection models for medical imaging. Future research should focus on integrating hybrid detection architectures, optimizing training strategies, and leveraging advanced augmentation techniques \cite{b24} to improve robustness in real-world clinical settings. 







\onecolumn
\appendix
\section{Architectural Overviews of Models} \label{app:models}

This appendix provides architectural diagrams of the DINO-DETR, Grounding DINO, and YOLO models. The diagrams are adapted from \cite{b13}.

\subsection{DINO-DETR}
\begin{figure}[htbp]
    \centering
    \includegraphics[width=1\linewidth]{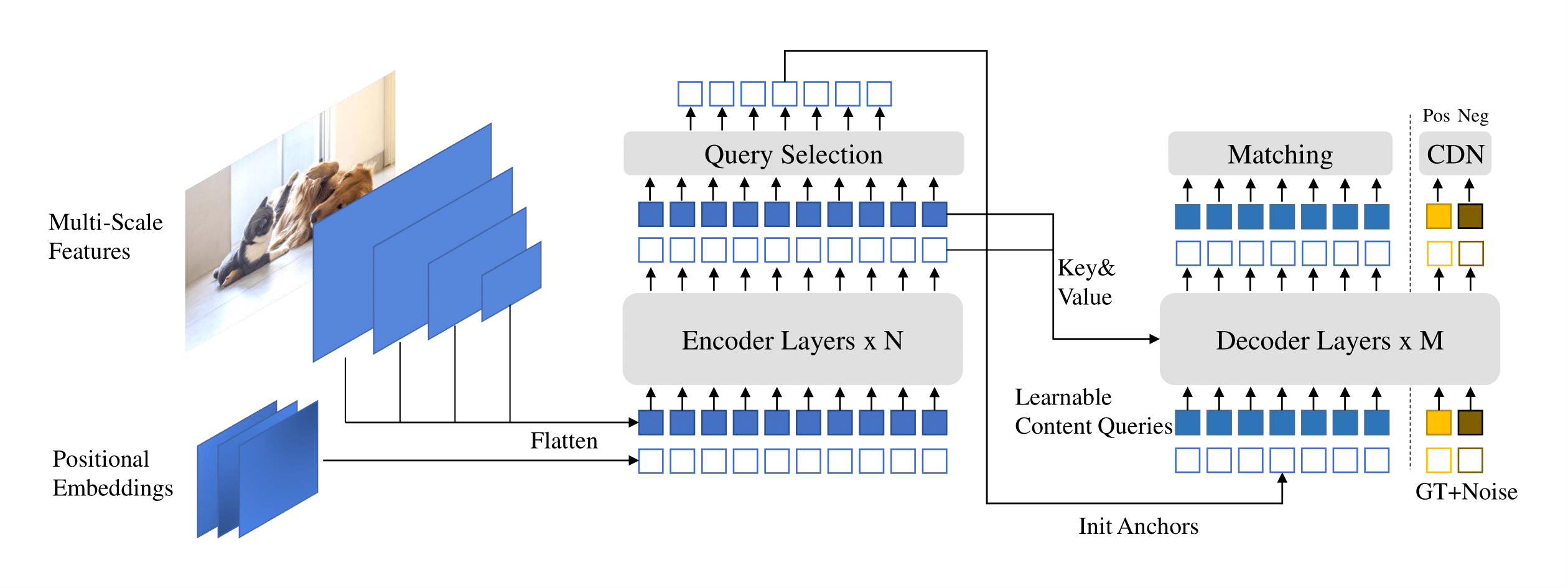} 
    \caption{Architectural diagram of DINO-DETR, adapted from \cite{b6}.}
    \label{fig:dino-detr}
\end{figure}

\subsection{Grounding DINO}
\begin{figure}[htbp]
    \centering
    \includegraphics[width=1\linewidth]{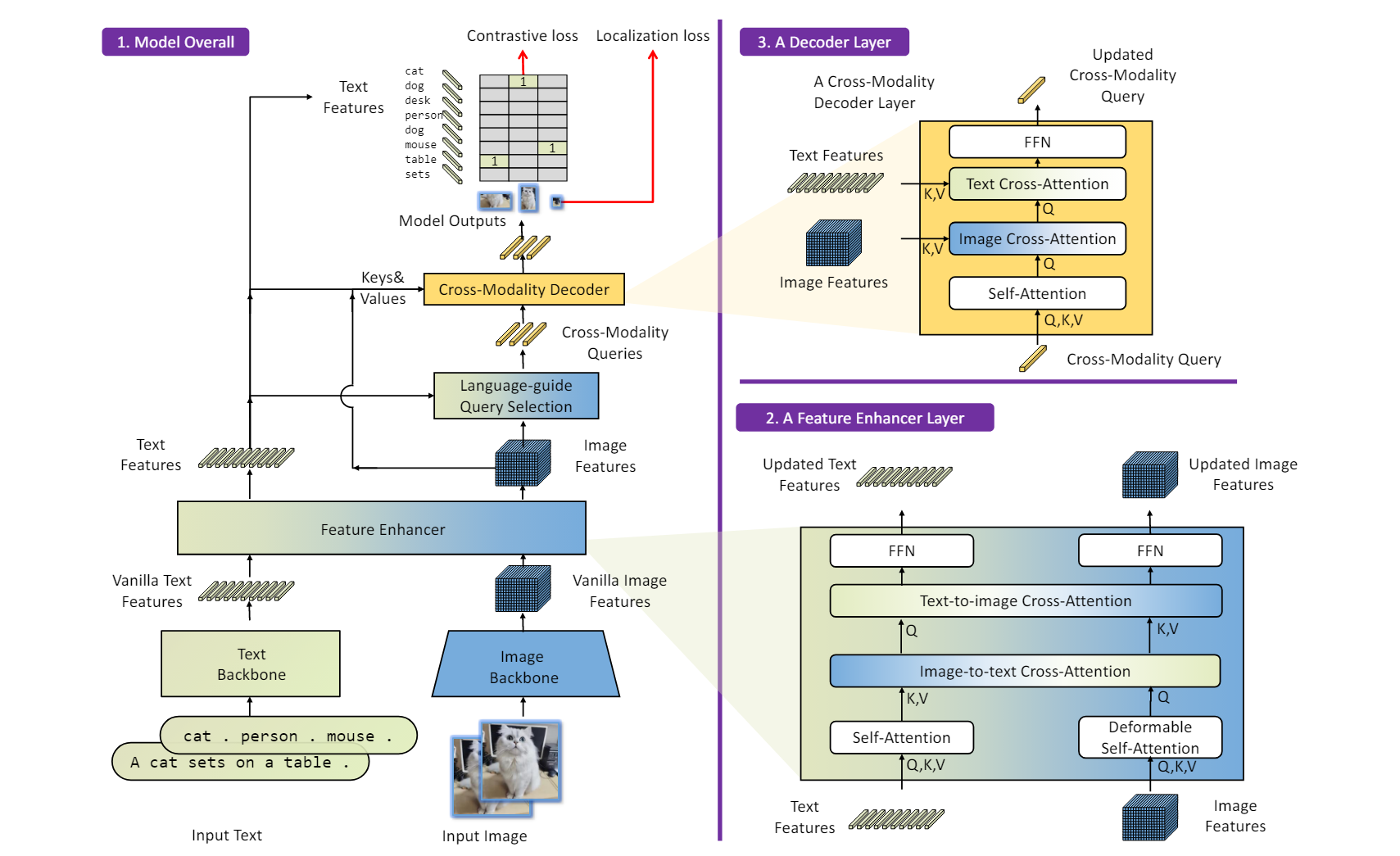} 
    \caption{Architectural diagram of Grounding DINO, adapted from \cite{b7}.}
    \label{fig:grounding-dino}
\end{figure}

\subsection{YOLO}
\begin{figure}[htbp]
    \centering
    \includegraphics[width=1\linewidth]{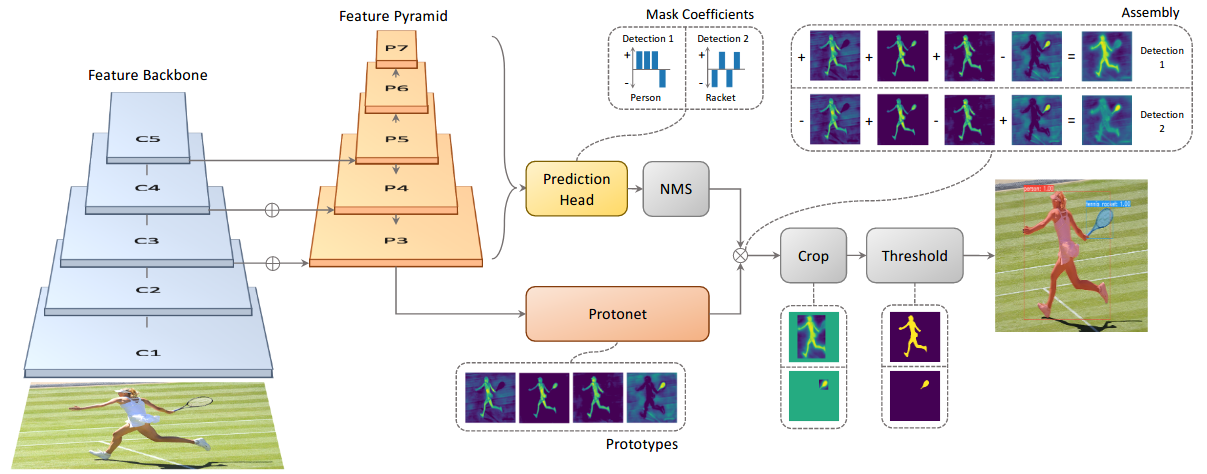} 
    \caption{Architectural diagram of YOLO, adapted from \cite{b5}.}
    \label{fig:yolo}
\end{figure}

\end{document}